# Robust Hypothesis Testing Using Wasserstein Uncertainty Sets

**Rui Gao**
School of Industrial and Systems Engineering
Georgia Institute of Technology
Atlanta, GA 30332
rgao32@gatech.edu

**Liyan Xie**
School of Industrial and Systems Engineering
Georgia Institute of Technology
Atlanta, GA 30332
lxie49@gatech.edu

**Yao Xie**
School of Industrial and Systems Engineering
Georgia Institute of Technology
Atlanta, GA 30332
yao.xie@isye.gatech.edu

**Huan Xu**
School of Industrial and Systems Engineering
Georgia Institute of Technology
Atlanta, GA 30332
huan.xu@isye.gatech.edu

## Abstract

We develop a novel computationally efficient and general framework for robust hypothesis testing. The new framework features a new way to construct uncertainty sets under the null and the alternative distributions, which are sets centered around the empirical distribution defined via Wasserstein metric, thus our approach is data-driven and free of distributional assumptions. We develop a convex safe approximation of the minimax formulation and show that such approximation renders a nearly-optimal detector among the family of all possible tests. By exploiting the structure of the least favorable distribution, we also develop a tractable reformulation of such approximation, with complexity independent of the dimension of observation space and can be nearly sample-size-independent in general. Real-data example using human activity data demonstrated the excellent performance of the new robust detector.

## 1 Introduction

Hypothesis testing is a fundamental problem in statistics and an essential building block for scientific discovery and many machine learning problems such as anomaly detection. The goal is to develop a *decision rule*, also called the *detector*, which can discriminate between two (or multiple) hypotheses based on data and achieve small error probability. For simple hypothesis test, it is well-known from the Neyman-Pearson Lemma that the likelihood ratio between two hypotheses is optimal. However, in practice, when the true distribution deviates from the assumed nominal distribution, especially in the presence of outliers, the performance of the likelihood ratio detector is no longer optimal and it may perform poorly.

Various robust detectors have been developed, to address the issue with distribution misspecification and outliers. The robust detectors are constructed by introducing various uncertainty sets for the distributions under the null and the alternative hypotheses. Huber's original work [12] considers the so-called $\epsilon$-contamination sets, which contain distributions that are close to the nominal distributions in terms of total variation metric. The more recent works [16, 9] consider uncertainty set induced by Kullback-Leibler divergence around a nominal distribution. Based on this, robust detectors usually depend on the so-called least-favorable distributions (LFD). Although there has been much success in theoretical results, *computation* remains a major challenge in finding robust detectors and finding LFD in general. Existing results are usually only for the one-dimensional setting. In multi-dimensional setting, finding LFD remains an open question in the literature.

In this paper, we present a novel computationally efficient framework for developing robust minimax detectors. We develop a nearly-optimal convex approximation of the minimax formulation and show that the optimal detector and the LFD can be solved efficiently by convex optimization by exploiting the solution structure.

A notable difference of our framework from prior work is a new way to construct uncertainty sets: we use *empirical distributions* as the nominal distributions and the sets are defined using Wasserstein metrics. The motivation for considering *new constructions for uncertainty sets* is three-fold:

- By using empirical distribution to define uncertainty sets, we enable *data-driven* approach. This is quite general since we do not assume any parametric forms for the distributions, but rather "let the data speak for itself."
- Wasserstein metric is a more *flexible* measure of closeness between two distributions comparing to divergence measures including Kullback-Leibler divergence. The latter is not well-defined for distributions with non-overlapping support, which occurs often in (1) *data-driven* problems, in which we often want to measure the closeness between an empirical distribution and some continuous underlying true distribution, and (2) *high-dimensional* problems, in which we may want to compare two distributions that are of high dimensions, but supported on two low-dimension manifolds with measure-zero intersection.
- It often renders a *computationally efficient* reformulation of the minimax robust hypothesis testing problem.

Hence, our approach is particularly good when the number of observations is small relative to the dimension of the observation space, when parametric approach becomes difficulty (as it is hard to estimate the parameters, e.g., estimating the mean and covariance matrix of high-dimensional Gaussian).

To solve the minimax robust detector problem, we face at least two difficulties: (i) The optimization over the all possible detectors is hard in general since we consider any infinite-dimensional detector with nonlinear dependence on data; (ii) The worst-case distribution over the uncertainty sets is also an infinite dimensional optimization problems in general. To tackle these difficulties, in Section 3, we develop a safe approximation of the minimax formulation by considering a family of tests with a special form that facilitates a convex approximation. We show that such approximation renders a nearly-optimal detector among the family of all possible tests (Theorem 1), and the risk of the optimal detector is closely related to divergence measures (Theorem 2). In Section 4, exploiting the structure of the least favorable distributions, we derive a finite-dimensional convex programming reformulation of the safe approximation based on strong duality (Theorem 3). Finally, Section 5 demonstrates the excellent performance of our robust detectors using real-data for human activity detection.

## 2 Problem Set-up and Related Work

Let $\Omega \subset \mathbb{R}^d$ be the observation space where the observed random variable takes its values. Denote by $\mathscr{P}(\Omega)$ be the set of all probability distributions on $\Omega$. Let $\mathcal{P}_1, \mathcal{P}_2 \subset \mathscr{P}(\Omega)$ be our uncertainty sets associated with hypothesis $H_1$ and $H_2$. The uncertainty sets are two families of probability distributions on $\Omega$. We assume that the true probability distribution of the observed random variable belongs to either $\mathcal{P}_1$ or $\mathcal{P}_2$. Given an observation $\omega$ of the random variable, we would like to decide which one of the following hypotheses is true

$$\begin{aligned} H_1 &: \quad \omega \sim P_1, \quad P_1 \in \mathcal{P}_1, \\ H_2 &: \quad \omega \sim P_2, \quad P_2 \in \mathcal{P}_2. \end{aligned}$$

A *test* for this testing problem is a (Lebesgue) measurable function $T : \Omega \to \{1, 2\}$. Given an observation $\omega \in \Omega$, the test accepts hypotheses $H_{T(\omega)}$ and rejects the other. A test is called *simple*, if $\mathcal{P}_1, \mathcal{P}_2$ are singletons.

The *worst-case risk of a test* is defined as the maximum of the worst-case type-I and type-II errors

$$\epsilon(T|\mathcal{P}_1, \mathcal{P}_2) := \max\Big( \sup_{P_1 \in \mathcal{P}_1} P_1\{\omega : T(\omega) = 1\}, \sup_{P_2 \in \mathcal{P}_1} P_2\{\omega : T(\omega) = 0\} \Big).$$

Here, without loss of generality, we define the risk to be the maximum of the two types of errors. Our framework can extend directly to the case where the risk is defined as a linear combination of the Type-I and Type-II errors (as usually considered in statistics).

We consider the *minimax robust hypothesis test* formulation, where the goal is to find a test that minimizes the worst-case risk. More specifically, given $\mathcal{P}_1, \mathcal{P}_2$ and $\epsilon > 0$, we would like to find an $\epsilon$-optimal solution of the following problem

$$\inf_{T:\Omega\to\{0,1\}} \epsilon(T|\mathcal{P}_1, \mathcal{P}_2). \tag{1}$$

We construct our uncertainty sets $\mathcal{P}_1, \mathcal{P}_2$ to be centered around two empirical distributions and defined using the Wasserstein metric. Given two empirical distributions $Q_k = (1/n_k)\sum_{i=1}^{n_k} \delta_{\widehat{\omega}_i^k}$, which are based on samples drawn from two underlying distributions respectively, where $\delta_\omega$ denotes the Dirac measure on $\omega$. Define the sets using Wasserstein metric (of order 1):

$$\mathcal{P}_k = \{P \in \mathscr{P}(\Omega) : \mathcal{W}(P, Q_k) \le \theta_k\},\ k = 1, 2, \tag{2}$$

where $\theta_k > 0$ specifies the radius of the set, and $\mathcal{W}(P, Q)$ denotes the Wasserstein metric of order 1:

$$\mathcal{W}(P, Q) := \min_{\gamma\in\mathscr{P}(\Omega^2)} \left\{\mathbb{E}_{(\omega,\omega')\sim\gamma}\left[\|\omega - \omega'\|\right] :\ \gamma \text{ has mariginal distributions } P \text{ and } Q\right\},$$

where $\|\cdot - \cdot\|$ is an arbitrary norm on $\mathbb{R}^n$. We consider Wasserstein metric of order 1 for the ease of exposition. Intuitively, the joint distribution $\gamma$ on the right-hand side of the equation above can be viewed as a transportation plan which transports probability mass from $P$ to $Q$. Thus, the Wasserstein metric between two distributions equals the cheapest cost (measured in some norm $\|\cdot - \cdot\|$) of transporting probability mass from one distribution to the other. In particular, if both $P$ and $Q$ are finite-supported, the above minimization problem reduces to the transportation problem in linear programming. Wasserstein metric has recently become popular in machine learning as a way to measuring the distance between probability distributions, and has been applied to a variety of areas including computer vision [23, 15, 21], generative adversarial networks [2, 10], and distributionally robust optimization [6, 7, 4, 26, 24].

### 2.1 Related Work

We present a brief review on robust hypothesis test and related work. The most commonly seen form of hypothesis test in statistics is simple hypothesis. The so-called simple hypothesis test assuming that the null and the alternative distributions are two singleton sets. Suppose one is interested in discriminating between $H_0 : \theta = \theta_0$ and $H_1 : \theta = \theta_1$, when the data $x$ is assumed to follow a distribution $f_\theta$ with parameter $\theta$. The likelihood ratio test rejects $H_0$ when $f_{\theta_1}(x)/f_{\theta_0}(x)$ exceeds a threshold. The celebrated Neyman-Pearson lemma says that the likelihood ratio is the most powerful test given a significance level. In other words, the likelihood ratio test achieves the minimum Type-II error given any Type-I error. In practice, when the true distributions deviate from the two assume distributions, especially in the presence of outliers, the likelihood ratio test is no longer optimal. The so-called robust detector aims to extend the simple hypothesis test to composite test, where the null and the alternative hypothesis includes a family of distributions. There are two main approaches to the minimax robust hypothesis testing, one dates back to Huber's seminal work [12], and one is attribute to [16]. Huber considers composite hypotheses over the so-called $\epsilon$-contamination sets which are defined as total variation classes of distributions around nominal distribution, where the more recent work [16, 9] considers uncertainty set defined using the Kullback-Leibler (KL) divergence, and demonstrated various closed-form LFD for one-dimensional setting. However, in the multi-dimensional setting, there remains the computational challenge to establish robust sequential detection procedures or to find the LFDs. Indeed, closed-form LFDs are found only for one-dimensional case (e.g, [11, 17, 9]) for one-dimensional case. Moreover, classic hypothesis test is usually parametric in that the distribution functions under the null and the alternative are assumed to be belong to family of distributions with certain parameters.

Recent works [8, 13] take a different approach from the classic statistical approach for hypothesis testing. Although "robust hypothesis test" are not mentioned, the formulation therein is essentially minimax robust hypothesis test, when the null and the alternative distributions parametric with the parameters belong to certain convex sets. They show that when exponential function is used as a convex relaxation, the optimal detector corresponds to the likelihood ratio test between the two LFD whose parameters are solved from a convex program. Our work is inspired by [8, 13] and our work extends the state-of-the-art in several ways. Convex relaxation of the 0-1 loss by the exponential function is considered in these prior work; we consider other *tighter* convex relaxations which may have smaller relaxation gap. We also show that, under these different convex relaxations, we can derive similar insights as [8] that the expected value of the worst-case loss will have a clear statistical meaning and corresponds to Hellinger distance, Jensen-Shannon divergence, etc. In addition, [8, 13] consider parametric distributions and uncertainty sets for parameters of these distributions and impose restrictive requirement on the considered families of distributions. In contrast, we consider a non-parametric uncertainty set defined using empirical distribution and Wasserstein metric, and impose no conditions on the considered distributions. We finally remark that [22] also considered using Wasserstein metric for hypothesis testing and drew connections between different test statistics. Our focus is different from theirs as we consider Wasserstein metric for the minimax robust formulation.

# 3 Optimal Detector

We consider a family of tests with a special form, which is referred as a *detector*. A detector $\phi : \Omega \to \mathbb{R}$ is a measurable function associated with a test $T_\phi$ which, for a given observation $\omega \in \Omega$, accepts $H_1$ and rejects $H_2$ whenever $\phi(\omega) \geq 0$, otherwise accepts $H_2$ and rejects $H_1$. The restriction of problem (1) on the sets of all detectors is

$$\inf_{\phi:\Omega\to\mathbb{R}} \max\Big( \sup_{P_1\in\mathcal{P}_1} P_1\{\omega : \phi(\omega) < 0\}, \sup_{P_2\in\mathcal{P}_1} P_2\{\omega : \phi(\omega) \geq 0\}\Big). \tag{3}$$

We next develop a safe approximation of problem 3 that provides an upper bound via convex approximations of the indicator function [20]. We introduce a notion called *generating function*.

**Definition 1** (Generating function)**.** A generating function $\ell : \mathbb{R} \to \mathbb{R}_+ \cup \{\infty\}$ is a nonnegative valued, nondecreasing, convex function satisfying $\ell(0) = 1$ and $\lim_{t\to-\infty} \ell(t) = 0$.

For any probability distribution $P$, it holds that $P\{\omega : \phi(\omega) < 0\} \leq \mathbb{E}_P[\ell(-\phi(\omega))\}]$. Set

$$\Phi(\phi; P_1, P_2) := \mathbb{E}_{P_1}[\ell \circ (-\phi)(\omega))] + \mathbb{E}_{P_2}[\ell \circ \phi(\omega)].$$

We define the *risk of a detector* for a test $(\mathcal{P}_1, \mathcal{P}_2)$ by

$$\epsilon(\phi|\mathcal{P}_1, \mathcal{P}_2) := \sup_{P_1\in\mathcal{P}_1, P_2\in\mathcal{P}_2} \Phi(\phi; P_1, P_2).$$

It follows that the following problem provides an upper bound of problem (3):

$$\inf_{\phi:\Omega\to\mathbb{R}} \sup_{P_1\in\mathcal{P}_1, P_2\in\mathcal{P}_2} \Phi(\phi; P_1, P_2). \tag{4}$$

We next bound the gap between (4) and (1). To facilitate discussion, we introduce an auxiliary function $\psi$, which is well-defined due to the assumptions on $\ell$:

$$\psi(p) := \min_{t\in\mathbb{R}} [p\ell(t) + (1-p)\ell(-t)], \quad 0 \leq p \leq 1.$$

For various generating functions $\ell$, $\psi$ admits a close-form expression. Table 1 lists some choices of generating function $\ell$ and their corresponding auxiliary function $\psi$. Note that the Hinge loss (last row in the table) leads to the smallest relaxation gap. As we shall see, $\psi$ plays an important role in our analysis, and is closely related to the divergence measure between probability distributions.

Table 1: Generating function (first column) and its corresponding auxiliary function (second column), optimal detector (third column), and detector risk (fourth column).

| $\ell(t)$ | $\psi(p)$ | $\phi^*$ | $1 - 1/2 \inf_\phi \Phi(\phi; P_1, P_2)$ |
|---|---|---|---|
| $\exp(t)$ | $2\sqrt{p(1-p)}$ | $\ln\sqrt{p_1/p_2}$ | $H^2(P_1, P_2)$ |
| $\log(1+\exp(t))/\log 2$ | $-(p\log p + (1-p)\log(1-p))/\log 2$ | $\log(p_1/p_2)$ | $JS(P_1, P_2)/\log 2$ |
| $(t+1)_+^2$ | $4p(1-p)$ | $1 - 2\frac{p_1}{p_1+p_2}$ | $\chi^2(P_1, P_2)$ |
| $(t+1)_+$ | $2\min(p, 1-p)$ | $\mathrm{sgn}(p_1 - p_2)$ | $TV(P_1, P_2)$ |

**Theorem 1** (Near-optimality of (4)). *Whenever there exists an feasible solution $T$ of problem (1) with objective value less than $\epsilon \in (0, 1/2)$, there exists a feasible solution $\phi$ of problem (4) with objective value less than $\psi(\epsilon)$.*

Theorem 1 guarantees that the approximation (4) of problem (1) is *nearly optimal*, in the sense that whenever there the hypotheses $H_1, H_2$ can be decided upon by a test $T$ with risk less than $\epsilon$, there exists a detector $\phi$ with risk less than $\psi(\epsilon)$. It holds regardless the specification of $\mathcal{P}_1$ and $\mathcal{P}_2$. Figure 1 illustrates the value of $\psi(\epsilon)$ when $\epsilon \in (0, 1/2)$.

The next proposition shows that we can interchange the inf and sup operators. Hence, in order to solve (4), we can first solve the problem of finding the best detector for a simple test $(P_1, P_2)$, and then finding the least favorable distribution that maximizes the risk among those best detectors.

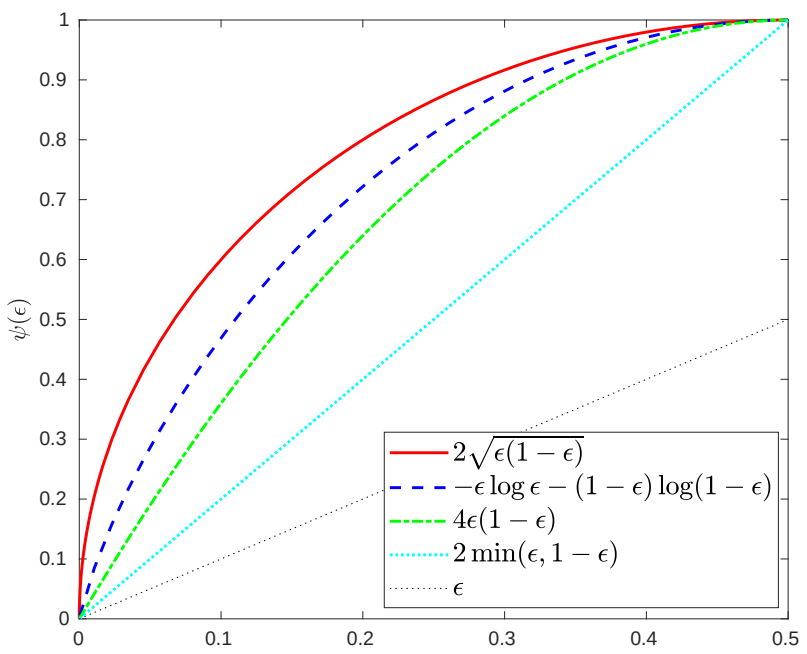

Figure 1: $\psi(\epsilon)$ as a function of $\epsilon$

**Proposition 1.** $\inf_{\phi:\Omega\to\mathbb{R}} \sup_{P_1\in\mathcal{P}_1, P_2\in\mathcal{P}_2} \Phi(\phi; P_1, P_2) = \sup_{P_1\in\mathcal{P}_1, P_2\in\mathcal{P}_2} \inf_{\phi:\Omega\to\mathbb{R}} \Phi(\phi; P_1, P_2).$

The next theorem provides an expression of the optimal detector and its risk.

**Theorem 2** (Optimal detector). *Let $\frac{dP_k}{d(P_1+P_2)}$ be the Radon-Nikodym derivative of $P_k$ with respect to $P_1 + P_2$, $k = 1, 2$. Then*

$$\inf_{\phi:\Omega\to\mathbb{R}} \Phi(\phi; P_1, P_2) = \int_\Omega \psi\big(\tfrac{dP_1}{d(P_1+P_2)}\big) d(P_1 + P_2).$$

*Define $\Omega_0(P_1, P_2) := \big\{\omega \in \Omega : 0 < \frac{dP_k}{d(P_1+P_2)}(\omega) < 1,\ k = 1, 2\big\}$. Suppose there exists a well-defined map $t : \Omega \to \mathbb{R}$ such that*

$$t^*(\omega) \in \arg\min_{t\in\mathbb{R}} \Big[\tfrac{dP_1}{d(P_1+P_2)}(\omega)\ell(-t) + \tfrac{dP_2}{d(P_1+P_2)}(\omega)\ell(t)\Big].$$

*Then $\phi^*(\omega) := -t^*(\omega)$ is an optimal detector for the simple test.*

**Remark 1.** By definition, $\psi(0) = \psi(1) = 0$. Then the infimum depends only on the value of $P_1, P_2$ on $\Omega_0$, the subset of $\Omega$ on which $P_1$ and $P_2$ are absolutely continuous with respect to each other:

$$\inf_{\phi:\Omega\to\mathbb{R}} \Phi(\phi; P_1, P_2) = \int_{\Omega_0} \psi\big(\tfrac{dP_1}{d(P_1+P_2)}\big) d(P_1 + P_2).$$

This is intuitive as we can always find a detector $\phi$ such that its value is arbitrarily close to zero on $\Omega \setminus \Omega_0$. In particular, if $P_1$ and $P_2$ have measure-zero overlap, then $\inf_\phi \Phi(\phi; P_1, P_2)$ equals zero, that is, the optimal test for the simple test $(P_1, P_2)$ has zero risk.

**Optimal detector** $\phi^*$. Set $p_k := dP_k/(d(P_1 + P_2))$ on $\Omega_0$, $k = 1, 2$. For the four choices of $\psi$ listed in Table 1, the optimal detectors $\phi^*$ on $\Omega_0$ are listed in the third column, where sgn denotes the sign function. The first one has been considered in [1].

**Relation between divergence measures and the risk of the optimal detector**. The term $\int_\Omega \psi\big(\frac{dP_1}{d(P_1+P_2)}\big) d(P_1 + P_2)$ can be viewed as a "measure of closeness" between probability distributions. Indeed, in the fourth column of Table 1 we show that the smallest detector risk for a simple test $P_1$ *vs.* $P_2$ equals the negative of some divergence between $P_1$ and $P_2$ up to a constant, where $H$, $JS$, $\Delta$, and $TV$ represent respectively the Hellinger distance, Jensen-Shannon divergence, triangle discrimination (symmetric $\chi^2$-divergence), and Total Variation metric [27]. It follows from Theorem 2 that

$$\sup_{P_1\in\mathcal{P}_1, P_2\in\mathcal{P}_2} \inf_{\phi:\Omega\to\mathbb{R}} \Phi(\phi; P_1, P_2) = \sup_{P_1\in\mathcal{P}_1, P_2\in\mathcal{P}_2} \int_\Omega \psi\big(\tfrac{dP_1}{d(P_1+P_2)}\big) d(P_1 + P_2). \tag{5}$$

The objective on the right-hand side is concave in $(P_1, P_2)$ since by Theorem 2, it equals the infimum of linear functions $\Phi(\phi; P_1, P_2)$ of $(P_1, P_2)$. Problem (5) can be interpreted as finding two distributions $P_1^* \in \mathcal{P}_1$ and $P_2^* \in \mathcal{P}_2$ such that the divergence between $P_1^*$ and $P_2^*$ is minimized. This makes sense in that the least favorable distribution $(P_1^*, P_2^*)$ should be as close to each other as possible for the worst-case hypothesis test scenario.

## 4 Tractable Reformulation

In this section, we provide a tractable reformulation of (5) by deriving a novel strong duality result. Recall in our setup, we are given two empirical distributions $Q_k = \frac{1}{n_k}\sum_{i=1}^{n_k} \delta_{\widehat{\omega}_k^i}$, $k = 1, 2$. To unify notation, for $l = 1, \ldots, n_1 + n_2$, we set

$$\omega^l = \begin{cases} \widehat{\omega}_1^l, & 1 \le l \le n_1, \\ \widehat{\omega}_2^{l-n_1}, & n_1 + 1 \le l \le n_1 + n_2, \end{cases}$$

and set $\widehat{\Omega} := \{\omega^l : \ l = 1, \ldots, n_1 + n_2\}$.

**Theorem 3** (Convex equivalent reformulation)**.** *Problem* (5) *with* $\mathcal{P}_1, \mathcal{P}_2$ *specified in* (2) *can be equivalently reformulated as a finite-dimensional convex program*

$$\begin{aligned} \max_{\substack{p_1, p_2 \in \mathbb{R}_+^{n_1+n_2} \\ \gamma_1, \gamma_2 \in \mathbb{R}_+^{(n_1+n_2)} \times \mathbb{R}_+^{(n_1+n_2)}}} \quad & \sum_{l=1}^{n_1+n_2} \psi\big(\tfrac{p_1^l}{p_1^l + p_2^l}\big)(p_1^l + p_2^l) \\ \text{subject to} \quad & \sum_{l=1}^{n_1+n_2} \sum_{m=1}^{n_1+n_2} \gamma_k^{lm} \left\| \omega^l - \omega^m \right\| \le \theta_k, \ k = 1, 2, \\ & \sum_m \gamma_1^{lm} = \frac{1}{n_1}, \ 1 \le l \le n_1, \quad \sum_m \gamma_1^{lm} = 0, \ n_1 + 1 \le l \le n_1 + n_2, \\ & \sum_m \gamma_2^{lm} = 0, \ 1 \le l \le n_1, \quad \sum_m \gamma_2^{lm} = \frac{1}{n_2}, \ n_1 + 1 \le l \le n_1 + n_2, \\ & \sum_l \gamma_k^{lm} = p_k^m, \ 1 \le m \le n_1 + n_2, \ k = 1, 2. \end{aligned} \tag{6}$$

Theorem 3, combining with Proposition 1, indicates that problem (4) is equivalent to problem (6).

**Decision variables**. $p_k$ can be identified with a probability distribution on $\widehat{\Omega}$, because $\sum_l p_k^l = \sum_{lm} \gamma_k^{lm} = 1$, and $\gamma_k$ can be viewed as a joint probability distribution on $\widehat{\Omega}^2$ with marginal distributions $Q_k$ and $p_k$. We can eliminate variable $p_1, p_2$ by substituting $p_k$ with $\gamma_k$ using the last constraint, so that $\gamma_1, \gamma_2$ are the only decision variables.

**Objective**. The objective function is identical to the objective function of (5), and thus we are maximizing a concave function of $(p_1, p_2)$. If we substitute $p_k$ with $\gamma_k$, then the objective function is also concave in $(\gamma_1, \gamma_2)$.

**Constraints**. The constraints are all linear. Note that $\omega^l$ are parameters, but not decision variables, thus $\left\|\omega^l - \omega^m\right\|$ can be computed before solving the program. The constraints all together are equivalent to the Wasserstein metric constraints $\mathcal{W}(Q_k, p_k) \le \theta_k$.

**Strong duality**. Problem (6) is a restriction of problem (4) in the sense that they have the same objective but (4) restricts the feasible region to the subset of distributions that are supported on a subset $\widehat{\Omega}$. Nevertheless, Theorem 3 guarantees that the two problems has the same optimal value, because there exists a least favorable distribution supported on $\widehat{\Omega}$, as explained below.

**Intuition on the reformulation**.

We here provide insights on the structural properties of the least favorable distribution that explain why the reduction in Theorem 3 holds. The complete proof of Theorem 3 can be found in Appendix A. Suppose $Q_k = \delta_{\widehat{\omega}_k}$, $k = 1, 2$, $\Omega = \mathbb{R}^d$ and $\psi(p) = 2\sqrt{p(1-p)}$. Note that Wasserstein metric measures the cheapest cost (measured in $\|\cdot - \cdot\|$) of transporting probability mass from one distribution to the other. Thus, based on the discussion in Section 3, the goal of problem (5) is to move (part of) the probability mass on $\omega_1$ and $\omega_2$ such that the negative divergence between the resulting distributions is maximized. The following three key observations demonstrate *how to move the probability mass in a least favorable way.*

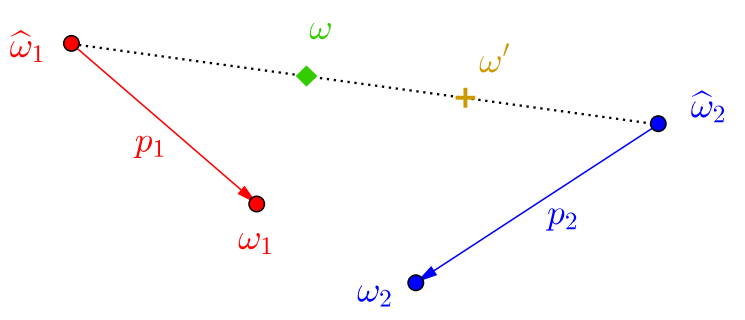

Figure 2: Illustration of the least favorable distribution: it is always better off to move the probability mass from $\widehat{\omega}_1$ and $\widehat{\omega}_2$ to an identical point $\omega$ on the line segment connecting $\widehat{\omega}_1, \widehat{\omega}_2$.

(i) Consider feasible solutions of the form

$$(P_1, P_2) = \big((1-p_1)\delta_{\widehat{\omega}_1} + p_1\delta_{\omega_1},\ (1-p_2)\delta_{\widehat{\omega}_2} + p_2\delta_{\omega_2}\big),\ \ \omega_1, \omega_2 \in \Omega \setminus \{\widehat{\omega}_1, \widehat{\omega}_2\}.$$

Namely, $(P_1, P_2)$ is obtained by moving out probability mass $p_k > 0$ from $\widehat{\omega}_k$ to $\widehat{\omega}_k$, $k = 1, 2$ (see Figure 2). It follows that the objective value

$$\int_\Omega \psi\left(\tfrac{dP_1}{d(P_1+P_2)}\right) d(P_1 + P_2) = \begin{cases} 2\sqrt{p_1 p_2}, & \text{if } \omega_1 = \omega_2, \\ 0, & o.w. \end{cases}$$

This is consistent with Remark 1 in that the objective value vanishes if the supports of $P_1, P_2$ do not overlap. Moreover, when $\omega_1 = \omega_2$, the objective value is independent of their common value $\omega = \omega_1 = \omega_2$. Therefore, we should move probability mass out of resources $\widehat{\omega}_1, \widehat{\omega}_2$ to some common region, which contain points that receive probability mass from both resources.

(ii) Motivated by (i), we consider solutions of the following form

$$(P_1, P_2) = \big((1-p_1)\delta_{\widehat{\omega}_1} + p_1\delta_{\omega},\ (1-p_2)\delta_{\widehat{\omega}_2} + p_2\delta_{\omega}\big),\ \ \omega \in \Omega \setminus \{\widehat{\omega}_1, \widehat{\omega}_2\},$$

which has the same objective value $2\sqrt{p_1 p_2}$. In order to save the budget for the Wasserstein metric constraint, i.e., to minimize the transport distance

$$p_1 \|\omega_1 - \widehat{\omega}_1\| + p_2 \|\omega_2 - \widehat{\omega}_2\|,$$

by triangle inequality we should choose $\omega_1 = \omega_2 = \omega$ to be on the line segment connecting $\widehat{\omega}_1$ and $\widehat{\omega}_2$ (see Figure 2).

(iii) Motivated by (ii), we consider solutions of the following form

$$P_k' = (1 - p_k - p_k')\delta_{\widehat{\omega}_k} + p_1\delta_{\omega_k} + p_1'\delta_{\omega_k'},\ k = 1, 2,$$

where $\omega_k, \omega_k' \notin \Omega \setminus \{\widehat{\omega}_k\}$ are on the line segment connecting $\widehat{\omega}_1$ and $\widehat{\omega}_2$, $k = 1, 2$. Then the objective value is maximized at $\omega_1 = \omega_1' = \widehat{\omega}_2$, $\omega_2 = \omega_2' = \widehat{\omega}_1$, and equals $2\sqrt{(p_1 + p_1')(p_2 + p_2')} + 2\sqrt{(1 - p_1 - p_1')(1 - p_2 - p_2')}$. Hence it is better off to move out probability mass from $\widehat{\omega}_1$ to $\widehat{\omega}_2$ and from $\widehat{\omega}_2$ to $\widehat{\omega}_1$.

Therefore, we conclude that there exist a least favorable distribution supported on $\widehat{\Omega}$. The argument above utilizes Theorem 2, the triangle inequality of a norm and the concavity of the auxiliary function $\psi$. The compete proof can be viewed as a generalization to the infinitesimal setting.

**Complexity**. Problem (6) is a convex program which maximizes a concave function subject to linear constraints. We briefly comment on the complexity of solving (6) in terms of the dimension of the observation space and the sample sizes:

(i) The complexity of (6) is *independent of the dimension* $d$ of $\Omega$, since we only need to compute pairwise distances $\|\omega^l - \omega^m\|$ as an input to the convex program.

(ii) The complexity in terms of the sample sizes $n_1, n_2$ depends on the objective function and can be *nearly sample size-independent* when the objective function is Lipschitz in $\ell_1$ norm (equivalently, the $\ell_\infty$ norm of the partial derivative is bounded). The reasons are as follows. In this case, after eliminating variables $p_1, p_2$, we end up with a convex program involving only $\gamma_1, \gamma_2$, and the Lipschitz constant of the objective with respect to $\gamma$ is identical to that with respect to $p$. Observe that the feasible region of each $\gamma_k$ is a subset of the $\ell_1$-ball in $\mathbb{R}_+^{(n_1+n_2)}$. Then according to the complexity theory of the first order method for convex optimization [3], when the objective function is Lipschitz in $\ell_1$ norm, the complexity is $O(\ln(n_1) + \ln(n_2))$. Notice that this is true for all except for the first case in Table 1. Hence, this is a quite general.

## 5 Numerical Experiments

In this section, we demonstrate the performance of our robust detector using real data for human activity detection. We adopt a dataset released by the Wireless Sensor Data Mining (WISDM) Lab in October 2013. The data in this set were collected with the Actitracker system, which is described in [18, 29, 14]. A large number of users carried an Android-based smartphone while performing various everyday activities. These subjects carried the Android phone in their pocket and were asked to walk, jog, ascend stairs, descend stairs, sit, and stand for specific periods of time.

The data collection was controlled by an application executed on the phone. This application is able to record the user's name, start and stop the data collection, and label the activity being performed. In all cases, the accelerometer data is collected every 50ms, so there are 20 samples per second. There are 2,980,765 recorded time-series in total. The activity recognition task involves mapping time-series accelerometer data to a single physical user activity [29]. Our goal is to detect the change of activity in real-time from sequential observations. Since it is hard to model distributions for various activities, traditional parametric methods do not work well in this case.

For each person, the recorded time-series contains the acceleration of the sensor in three directions. In this setting, every $\omega^l$ is a three-dimensional vector $(a_x^l, a_y^l, a_z^l)$. We set $\theta_1 = \theta_2 = \theta$ as the sample sizes are identical, and $\theta$ is chosen such that the quantity $1 - 1/2 \inf_\phi \Phi(\phi; P_1^*, P_2^*)$ in Table 1, or equivalently, the divergence between $P_1^*$ and $P_2^*$, is close to zero with high probability if $Q_1$ and $Q_2$ are bootstrapped from the data before change, where $P_1^*, P_1^*$ is the LFD yielding from (6). The intuition is that we want the Wasserstein ball to be large enough to avoid false detection while still have separable hypotheses (so the problem is well-defined).

We compare our robust detector, when coupled with CUSUM detector using a scheme similar to [5], with the Hotelling $\mathrm{T}^2$ control chart, which is a traditional way to detect the mean and covariance change for the multivariate case. The Hotelling control chart plots the following quantity [19]:

$$T^2 = (x - \mu)' \Sigma^{-1} (x - \mu),$$

where $\mu$ and $\Sigma$ are the sample mean and sample covariance obtained from training data.

As shown in Fig. 3 (a), in many cases, Hotelling $\mathrm{T}^2$ fails to detect the change successfully and our method performs pretty well. This is as expected since the change is hard to capture via mean and covariance as Hotelling does.

Moreover, we further test the proposed robust detector, $\phi^* = \frac{1}{2}\ln(p_1^*/p_2^*)$ and $\phi^* = \mathrm{sgn}(p_1^* - p_2^*)$, on 100 sequences of data. Here $p_1^*$ and $p_2^*$ are the LFD computed from the optimization problem (6). For each sequence, we choose the threshold for detection by controlling the type-I error. Then we compare the average detection delay of the robust detector and the Hotelling $\mathrm{T}^2$ control chart, as shown in 3 (b). The robust detector has a clear advantage, and the $\mathrm{sgn}(p_1^* - p_2^*)$ indeed has better performance than $\frac{1}{2}\ln(p_1^*/p_2^*)$, consistent with our theoretical finding.

## 6 Conclusion

In this paper, we propose a data-driven, distribution-free framework for robust hypothesis testing based on Wasserstein metric. We develop a computationally efficient reformulation of the minimax problem which renders a nearly-optimal detector. The framework is readily extended to multiple hypotheses and sequential settings. The approach can also be extended to other settings, such as constraining the Type-I error to be below certain threshold (as the typical statistical test of choosing the size or significance level of the test), or considering minimizing a weighed combination of the Type-I and Type-II errors. In the future, we will study the optimal selection of the size of the uncertainty sets leveraging tools from distributionally robust optimization, and test the performance of our framework on large-scale instances.

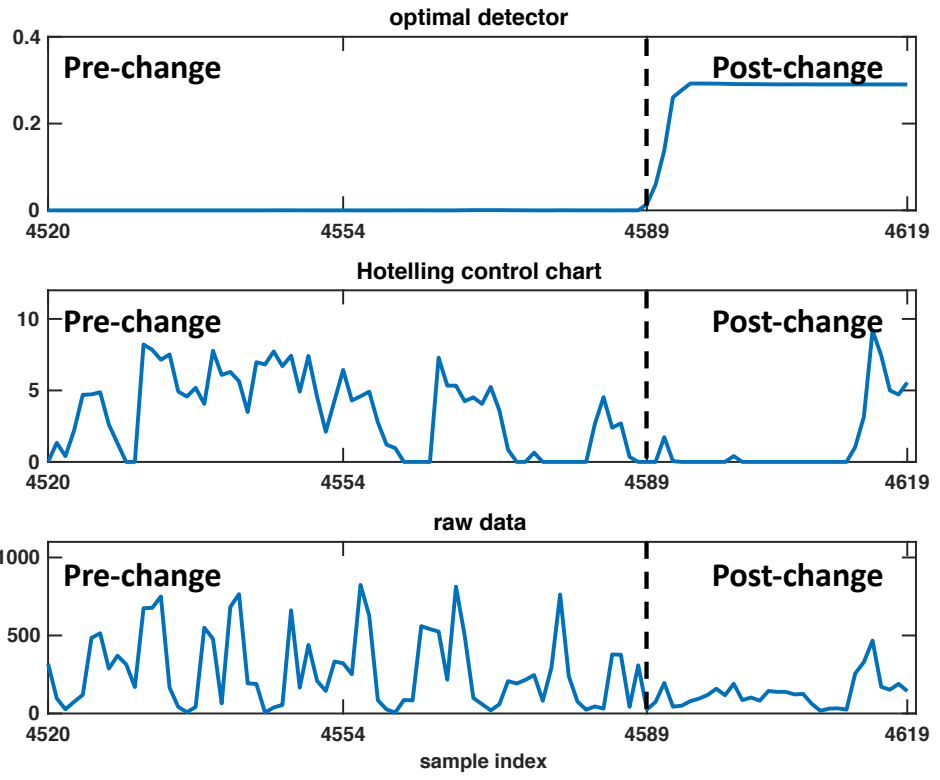

(a)

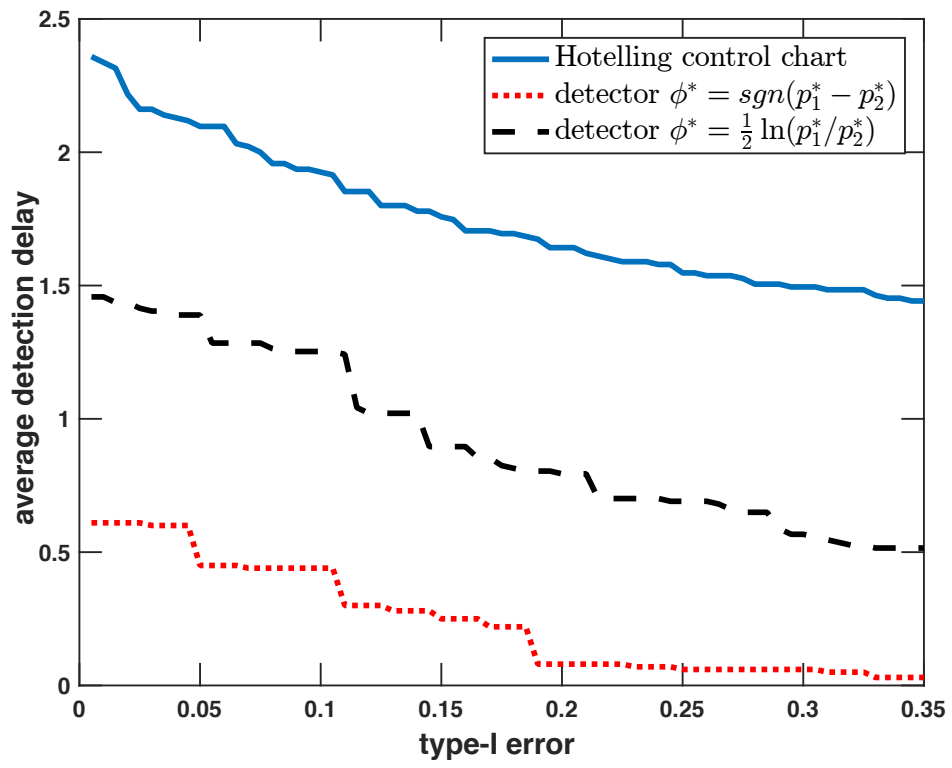

(b)

Figure 3: Comparison of the detector $\phi^* = \frac{1}{2}\ln(p_1^*/p_2^*)$ with Hotelling control chart: (a): Upper: the proposed optimal detector; Middle: the Hotelling $\mathrm{T}^2$ control chart; Lower: the raw data, here we plot $(a_x^2 + a_y^2 + a_z^2)^{1/2}$ for simple illustration. The dataset is a portion of full observations from the person indexed by 1679, with the pre-change activity jogging and post-change activity walking. The black dotted line at index 4589 indicates the boundary between the pre-change and post-change regimes. (b): The average detection delay v.s. type-I error. The average is taken over 100 sequences of data.

# A Proofs

*Proof of Theorem 1.* Let $T$ be a test with risk $\epsilon(T|\mathscr{P}_1, \mathscr{P}_2) \leq \epsilon$. For $\chi \in \{1, 2\}$, set $\Omega_\chi = \{\omega \in \Omega : T(\omega) = \chi\}$. Then $P_1(\Omega_2) \leq \epsilon$ for any $P_1 \in \mathscr{P}_1$, and $P_2(\Omega_1) \leq \epsilon$ for any $P_2 \in \mathscr{P}_2$. We choose $\phi$ such that

$$\phi(\omega) = \begin{cases} c_1, & \omega \in \Omega_1, \\ c_2, & \omega \in \Omega_2, \end{cases}$$

where $c_1 \geq 0 > c_2$. It follows that

$$\begin{aligned}
&\mathbb{E}_{P_1}[\ell \circ (-\phi)(\omega)] + \mathbb{E}_{P_2}[\ell \circ \phi(\omega)] \\
&= \mathbb{E}_{P_1}\left[\ell \circ (-\phi)(\omega) \cdot \mathbb{1}_{\Omega_1}(\omega)\right] + \mathbb{E}_{P_1}\left[\ell \circ (-\phi)(\omega) \cdot \mathbb{1}_{\Omega_2}(\omega)\right] \\
&\quad + \mathbb{E}_{P_2}\left[\ell \circ \phi(\omega) \cdot \mathbb{1}_{\Omega_1}(\omega)\right] + \mathbb{E}_{P_2}\left[\ell \circ \phi(\omega) \cdot \mathbb{1}_{\Omega_2}(\omega)\right] \\
&= \ell(-c_1) \cdot (1 - P_1(\Omega_2)) + \ell(-c_2) \cdot P_1(\Omega_2) \\
&\quad + \ell(c_1) \cdot P_2(\Omega_1) + \ell(c_2) \cdot (1 - P_2(\Omega_1)) \\
&\leq \ell(-c_1) + \epsilon \cdot (\ell(-c_2) - \ell(-c_1)) + \ell(c_2) + \epsilon(\ell(c_1) - \ell(c_2)) \\
&= \epsilon\ell(c_1) + (1 - \epsilon)\ell(-c_1) + \epsilon\ell(-c_2) + (1 - \epsilon)\ell(c_2).
\end{aligned}$$

where the inequality follows from $c_2 < c_1$ and the monotonicity of $\ell$. Since $\ell(c_1) > \ell(-c_1)$, $\ell(-c_2) > \ell(c_2)$, and $\epsilon \in (0, \frac{1}{2})$, minimizing over $c_1 \geq 0$ and $c_2 < 0$ yields the result. □

*Proof of Theorem 2.* Note that $P_1$, $P_2$ are absolutely continuous with respect to $P_1 + P_2$. We have that

$$\begin{aligned}
&\inf_{\phi:\Omega\to\mathbb{R}} \Phi(\phi; P_1, P_2) \\
&= \inf_{\phi:\Omega\to\mathbb{R}} \int_\Omega \left[\ell(-\phi(\omega))\frac{dP_1}{d(P_1+P_2)}(\omega) + \ell(\phi(\omega))\frac{dP_2}{d(P_1+P_2)}(\omega)\right] d(P_1 + P_2)(\omega)
\end{aligned}$$

For $\omega$ such that $\frac{dP_1}{d(P_1+P_2)}(\omega) = 1$, by assumptions on the generating function $\ell$, we can define $\phi$ such that $\ell(-\phi(\omega)) \leq \epsilon$ for an arbitrary $\epsilon > 0$. Similarly, for $\omega$ such that $\frac{dP_1}{d(P_1+P_2)}(\omega) = 0$, by assumptions on the generating function $\ell$, we can define $\phi$ such that $\ell(\phi(\omega)) \leq \epsilon$ for an arbitrary $\epsilon > 0$. Hence $\inf_\phi \Phi(\phi; \mu, \nu)$ depends on the integration only on $\Omega_0(P_1, P_2)$. It follows that

$$\begin{aligned}
&\inf_{\phi:\Omega\to\mathbb{R}} \Phi(\phi; P_1, P_2) \\
&= \inf_{\phi:\Omega\to\mathbb{R}} \int_{\Omega_0} \left[\ell(-\phi(\omega))\frac{dP_1}{d(P_1+P_2)}(\omega) + \ell(\phi(\omega))\frac{dP_2}{d(P_1+P_2)}(\omega)\right] d(P_1 + P_2)(\omega) \\
&= \int_{\Omega_0} \inf_{t\in\mathbb{R}} \left[\ell(-t)\frac{dP_1}{d(P_1+P_2)}(\omega) + \ell(t)\frac{dP_2}{d(P_1+P_2)}(\omega)\right] d(P_1 + P_2)(\omega),
\end{aligned}$$

where the last inequality follows by interchangeability principle [25]. □

*Proof of Theorem 3.*

*Step 1.* Using Lagrangian and Kantorovich's duality [28], we rewrite the problem as

$$
\begin{aligned}
&\sup_{\substack{P_1\in\mathcal{P}_1\\ P_2\in\mathcal{P}_2}} \int_\Omega \psi\big(\tfrac{dP_1}{d(P_1+P_2)}\big)d(P_1+P_2)\\
&= \sup_{P_1,P_2\in\mathscr{P}(\Omega)} \inf_{\lambda_1,\lambda_2\geq 0} \Bigg\{ \int_\Omega \psi\big(\tfrac{dP_1}{d(P_1+P_2)}\big)d(P_1+P_2) + \lambda_1\theta_1+\lambda_2\theta_2\\
&\quad - \sum_{k=1}^{2}\lambda_k \sup_{\substack{u_k\in\mathbb{R}^{n_k}\\ v_k\in L^1(P_k)}} \Big\{ \frac{1}{n_k}\sum_{i=1}^{n_k} u_k^i + \int_\Omega v_k dP_k : u_k^i + v_k(\omega) \leq \left\|\omega-\widehat{\omega}_k^i\right\|,\ \forall 1\leq i\leq n_k, \forall\omega\in\Omega\Big\}\Bigg\}\\
&= \sup_{P_1,P_2\in\mathcal{P}(\Omega)} \inf_{\substack{\lambda_1,\lambda_2\geq 0\\ u_k\in\mathbb{R}^{n_k}\\ v_k\in L^1(P_k)}} \Bigg\{ \int_\Omega \psi\big(\tfrac{dP_1}{d(P_1+P_2)}\big)d(P_1+P_2) + \lambda_1\theta_1+\lambda_2\theta_2\\
&\qquad - \sum_{k=1}^{2}\lambda_k\Big(\frac{1}{n_k}\sum_{i=1}^{n_k} u_k^i + \int_\Omega v_k dP_k\Big) : u_k^i + v_k(\omega)\leq \left\|\omega-\widehat{\omega}_k^i\right\|,\ \forall 1\leq i\leq n_k, \forall \omega\in\Omega\Bigg\}.
\end{aligned}
$$

Replacing $\lambda_k u_k^i$ with $u_k^i$ and $\lambda_k v_k$ with $v_k$, the second term and the constraints in the above equation are equivalently written as

$$
\sum_{k=1}^{2}\Big(\frac{1}{n_k}\sum_{i=1}^{n_k} u_k^i + \int_\Omega v_k dP_k\Big) : u_k^i + v_k(\omega)\leq \lambda_k\left\|\omega-\widehat{\omega}_k^i\right\|,\ \forall 1\leq i\leq n_k, \forall\omega\in\Omega.
$$

Note that such change of variable is valid even when $\lambda_k=0$. Further note that the objective function is non-increasing in $v_k$, thus we can replace $v_k$ with $\min_{1\leq i\leq n_k}[\lambda_k\left\|\omega-\widehat{\omega}_k^i\right\| - u_k^i]$ without changing the optimal value. Interchanging $\sup$ and $\inf$ yields

$$
\begin{aligned}
&\sup_{\substack{P_1\in\mathcal{P}_1\\ P_2\in\mathcal{P}_2}} \int_\Omega \psi\big(\tfrac{dP_1}{d(P_1+P_2)}\big)d(P_1+P_2)\\
&\leq \inf_{\lambda_1,\lambda_2,u_k,v_k} \Big\{\lambda_1\theta_1+\lambda_2\theta_2 - \sum_{k=1}^{2}\frac{1}{n_k}\sum_{i=1}^{n_k}u_k^i + \sup_{P_1,P_2\in\mathscr{P}(\Omega)} \int_\Omega \psi\big(\tfrac{dP_1}{d(P_1+P_2)}\big)d(P_1+P_2) - \int_\Omega\sum_{k=1}^{2} v_k dP_k\Big\},
\end{aligned}
\tag{7}
$$

where the inifimum is taken over the set

$$
\Big\{\lambda_1,\lambda_2\geq 0, u_k\in\mathbb{R}^{n_k}, v_k(\omega) = \min_{1\leq i\leq n_k}[\lambda_k\left\|\omega-\widehat{\omega}_k^i\right\| - u_k^i],\ \forall\omega\in\Omega,\ k=1,2\Big\}.
$$

*Step 2.* We next simplify the inner supremum in (7). We have

$$
\begin{aligned}
&\sup_{P_1,P_2\in\mathscr{P}(\Omega)} \Big\{ \int_\Omega \psi\big(\tfrac{dP_1}{d(P_1+P_2)}\big)d(P_1+P_2) - \sum_{k=1}^{2}\int_\Omega \min_{1\leq i\leq n_k}[\lambda_k\left\|\omega-\widehat{\omega}_k^i\right\| - u_k^i]\, dP_k\Big\}\\
&= \max_{\substack{1\leq i_k\leq n_k\\ k=1,2}} \sup_{P_1,P_2\in\mathscr{P}(\Omega)} \Big\{ \int_\Omega \Big(\psi\big(\tfrac{dP_1}{d(P_1+P_2)}(\omega)\big) - \sum_{k=1}^{2}\big[\lambda_k\left\|\omega-\widehat{\omega}_k^{i_k}\right\| - u_k^{i_k}\big]\tfrac{dP_k}{d(P_1+P_2)}(\omega)\Big)d(P_1+P_2)(\omega)\Big\}
\end{aligned}
$$

For a given solution $(P_1,P_2)$ and for $\omega\in\operatorname{supp}(P_1+P_2)$, set

$$
T(\omega) = \arg\min_{\omega'\in\Omega}\Big\{\sum_{k=1}^{2}\big[\lambda_k\left\|\omega'-\widehat{\omega}_k^{i_k}\right\| - u_k^{i_k}\big]\tfrac{dP_k}{d(P_1+P_2)}(\omega)\Big\} = \begin{cases}\widehat{\omega}_1^{i_1}, & \text{if } \lambda_1\tfrac{dP_1}{d(P_1+P_2)}(\omega)\geq \lambda_2\tfrac{dP_2}{d(P_1+P_2)}(\omega),\\ \widehat{\omega}_2^{i_2}, & o.w.\end{cases}
$$

Define another solution $(P_1',P_2')$ such that $P_k'(B) = P_k\{\omega\in\Omega : T(\omega)\in B\}$ for any measurable set $B\subset\Omega$. Then by definition of $T$ we have that

$$
\sum_{k=1}^{2}\int_\Omega \lambda_k\left\|\omega_k-\omega\right\| dP_k'(\omega) \leq \sum_{k=1}^{2}\int_\Omega\lambda_k\left\|\omega_k-\omega\right\| dP_k(\omega).
$$

In addition, by concavity of $\psi$, for any $p_1, p_1', p_2, p_2' > 0$, it holds that

$$\psi(\tfrac{p_1}{p_1+p_2})(p_1+p_2) + \psi(\tfrac{p_1'}{p_1'+p_2'})(p_1'+p_2') \leq \psi(\tfrac{p_1+p_1'}{p_1+p_2+p_1'+p_2'})(p_1+p_2+p_1'+p_2'),$$

thus it follows that

$$\int_\Omega \psi\big(\tfrac{dP_1}{d(P_1+P_2)}\big)d(P_1+P_2) \leq \int_\Omega \psi\big(\tfrac{dP_1'}{d(P_1'+P_2')}\big)d(P_1'+P_2').$$

Hence $(P_1', P_2')$ is a feasible solution that yields an objective value no worse than $(P_1, P_2)$. This suggests that in order to solve the inner supremum of (7), it suffices to only consider $(P_1, P_2)$ such that $\operatorname{supp} P_k \subset \hat{\Omega} := \operatorname{supp} Q_1 \cup \operatorname{supp} Q_2$.

For $l = 1, \ldots, n_1+n_2$, set $p_l^k = P_k(\omega^l)$, and note that $\gamma_k \in \Gamma(P_k, Q_k)$ can be identified with a non-negative matrix $\gamma_k \in \mathbb{R}_+^{n_1+n_2} \times \mathbb{R}_+^{n_1+n_2}$ with all the column and row sums being 1. Thus, the inner supremum in (7) can now be equivalently written as

$$\sup_{\substack{p_1,p_2\in\mathbb{R}_+^{n_1+n_2}\\ \sum_l p_k^l=1}} \Big\{ \sum_{1\leq l\leq n_1+n_2} \psi\big(\tfrac{p_1^l}{p_1^l+p_2^l}\big)(p_1^l+p_2^l) - \sum_{k=1}^{2}\sum_{l=1}^{n_1+n_2} p_k^l \min_{1\leq i\leq n_k} \cdot[\lambda_k \left\|\omega^l - \widehat{\omega}_k^i\right\| - u_k^i] \Big\}$$

*Step 3*. It follows from Step 2 that

$$\begin{aligned}
&\sup_{\substack{P_1\in\mathcal{P}_1\\ P_2\in\mathcal{P}_2}} \int_\Omega \psi\big(\tfrac{dP_1}{d(P_1+P_2)}\big)d(P_1+P_2)\\
&\leq \inf_{\lambda_1,\lambda_2\geq 0} \Big\{\lambda_1\theta_1 + \lambda_2\theta_2 + \sup_{\substack{p_1,p_2\in\mathbb{R}_+^{n_1+n_2}\\ \sum_l p_k^l=1}} \Big\{ \sum_{1\leq l\leq n_1+n_2} \psi\big(\tfrac{p_1^l}{p_1^l+p_2^l}\big)(p_1^l+p_2^l)\\
&\qquad - \sum_{k=1}^{2}\sum_{l=1}^{n_1+n_2} p_k^l \min_{1\leq i\leq n_k} \cdot[\lambda_k \left\|\omega^l - \widehat{\omega}_k^i\right\| - u_k^i] \Big\}\Big\}.
\end{aligned}$$

Applying finite-dimensional convex programming duality on the right-hand side and reverse the procedure in Step 1, we obtain that the right-hand side of (7) is equivalent to (6). Observe that both sides of (7) have the same objective function, but the feasible region of the right-hand side is a subset of that of the left-hand side, and thus the right-hand side should be no greater than the left-hand side, i.e., the above inequality should hold as equality. Thereby we complete the proof.

□

*Proof of Proposition 1*. It suffices to show that

$$\inf_{\phi:\Omega\to\mathbb{R}} \sup_{P_1\in\mathcal{P}_1, P_2\in\mathcal{P}_2} \Phi(\phi; P_1, P_2) \leq \sup_{P_1\in\mathcal{P}_1, P_2\in\mathcal{P}_2} \int_\Omega \psi\big(\tfrac{dP_1}{d(P_1+P_2)}\big)d(P_1+P_2).$$

Using the strong duality result for distributionally robust optimization with Wasserstein metric [7], $\sup_{P_1\in\mathcal{P}_1,P_2\in\mathcal{P}_2}\Phi(\phi;P_1,P_2)$ has an equivalent dual formulation

$$\begin{aligned}
&\sup_{P_1\in\mathcal{P}_1,P_2\in\mathcal{P}_2}\Phi(\phi;P_1,P_2)\\
&=\inf_{\substack{\lambda_1,\lambda_2\geq 0\\ \pi^1_{\#}\gamma_k=Q_k}}\Big\{\lambda_1\theta_1+\lambda_2\theta_2+\int_{\Omega^2}\big[\ell(-\phi(\omega_1))-\lambda_1\|\widehat{\omega}_1-\omega_1\|\big]\gamma_1(d\widehat{\omega}_1,d\omega_1)\\
&\qquad\qquad+\int_{\Omega^2}\big[\ell(\phi(\omega_2))-\lambda_2\|\widehat{\omega}_2-\omega_2\|\big]\gamma_2(d\widehat{\omega}_2,d\omega_2)\Big\}\\
&\leq\lambda_1^*\theta_1+\lambda_2^*\theta_2+\int_{\Omega^2}\big[\ell(-\phi(\omega_1))-\lambda_1^*\|\widehat{\omega}_1-\omega_1\|\big]\gamma_1^*(d\widehat{\omega}_1,d\omega_1)\\
&\qquad\qquad+\int_{\Omega^2}\big[\ell(\phi(\omega_2))-\lambda_2^*\|\widehat{\omega}_2-\omega_2\|\big]\gamma_2^*(d\widehat{\omega}_2,d\omega_2)\\
&=\lambda_1^*\theta_1+\lambda_2^*\theta_2+\int_{\Omega}\big[\ell(-\phi(\omega))\tfrac{d\pi^1_{\#}\gamma_1^*}{d(\pi^1_{\#}\gamma_1^*+\pi^1_{\#}\gamma_2^*)}(\omega)+\ell(\phi(\omega))\tfrac{d\pi^1_{\#}\gamma_2^*}{d(\pi^1_{\#}\gamma_1^*+\pi^1_{\#}\gamma_2^*)}(\omega)\big]d(\pi^1_{\#}\gamma_1^*+\pi^1_{\#}\gamma_2^*)(\omega)\\
&\qquad\qquad-\int_{\Omega^2}\lambda_1^*\|\widehat{\omega}_1-\omega_1\|\,\gamma_1^*(d\widehat{\omega}_1,d\omega_1)-\int_{\Omega^2}\lambda_2^*\|\widehat{\omega}_2-\omega_2\|\,\gamma_2^*(d\widehat{\omega}_2,d\omega_2),
\end{aligned}\tag{8}$$

where $(\lambda_1^*,\lambda_2^*)$ is dual optimizer of problem (7) associated with the first constraint, $(\gamma_1^*,\gamma_2^*)$ is the optimizer of (7), $\pi^1_{\#}\gamma_k$ denotes the first marginal distribution of $\gamma_k$, and $\pi^{jk}_{\#}\gamma$ denotes the marginal distribution of $\gamma$ projected on the $j$-th and the $k$-th coordinates.

On the other hand, by Theorem 3 it holds that

$$\begin{aligned}
&\sup_{P_1\in\mathcal{P}_1,P_2\in\mathcal{P}_2}\int_{\Omega}\psi\big(\tfrac{dP_1}{d(P_1+P_2)}\big)d(P_1+P_2)\\
&=\sup_{P_1\in\mathcal{P}_1,P_2\in\mathcal{P}_2}\int_{\Omega}\inf_{t\in\mathbb{R}}\Big[\ell(-t)\tfrac{dP_1}{d(P_1+P_2)}(\omega)+\ell(t)\tfrac{dP_2}{d(P_1+P_2)}(\omega)\Big]d(P_1+P_2)(\omega)\\
&=\lambda_1^*\theta_1+\lambda_2^*\theta_2+\int_{\Omega}\inf_{t\in\mathbb{R}}\big[\ell(-t)\tfrac{d\pi^1_{\#}\gamma_1^*}{d(\pi^1_{\#}\gamma_1^*+\pi^1_{\#}\gamma_2^*)}(\omega)+\ell(t)\tfrac{d\pi^1_{\#}\gamma_2^*}{d(\pi^1_{\#}\gamma_1^*+\pi^1_{\#}\gamma_2^*)}(\omega)\big]d(\pi^1_{\#}\gamma_1^*+\pi^1_{\#}\gamma_2^*)(\omega)\\
&\qquad\qquad-\int_{\Omega^2}\lambda_1^*\|\widehat{\omega}_1-\omega_1\|\,\gamma_1^*(d\widehat{\omega}_1,d\omega_1)-\int_{\Omega^2}\lambda_2^*\|\widehat{\omega}_2-\omega_2\|\,\gamma_2^*(d\widehat{\omega}_2,d\omega_2).
\end{aligned}$$

Comparing with (8), we have that for any $\epsilon>0$, there exists a detector $\phi_\epsilon$, such that

$$\sup_{P_1\in\mathcal{P}_1,P_2\in\mathcal{P}_2}\Phi(\phi_\epsilon;P_1,P_2)\leq\sup_{P_1\in\mathcal{P}_1,P_2\in\mathcal{P}_2}\int_{\Omega}\psi\big(\tfrac{dP_1}{d(P_1+P_2)}\big)d(P_1+P_2)+\epsilon.$$

Let $\epsilon\to 0$ yields the result.

□